\documentclass{llncs}
\usepackage[latin1]{inputenc}
\usepackage{amsmath}
\usepackage{amssymb}
\usepackage{latexsym,xspace}
\usepackage{color,float,url}
\usepackage{graphics,graphicx,float}
\usepackage{colortbl,graphics,graphicx}
\usepackage[ruled, linesnumbered,noend]{algorithm2e}

\usepackage{helvet}
\usepackage{courier}
\usepackage{multicol}
\usepackage{multirow}

\newcommand{\Rsats}{\texttt{Rsat+DS}\xspace}
\newcommand{\Minisats}{\texttt{Minisat+DS}\xspace}

\newcommand{\true}{\textsl{true}}
\newcommand{\Minisat}{\texttt{Minisat}\xspace}
\newcommand{\Rsat}{\texttt{Rsat}\xspace}

\newcommand{\false}{\textsl{false}}
\newcommand{\resolvent}{\sigma}

\newcommand{\classicalclause}{{imp}}

\newcommand{\textcanbeskipped}[1]{}

\begin{document}

\floatstyle{plain} 
\newfloat{Algorithm}{thp}{lop}
\floatname{Algorithm}{Algorithm}

\title{ 
 Learning for Dynamic Subsumption  
}

\author{Y.  Hamadi$^1$ \and S. Jabbour$^2$\and L. Sa\"is$^2$}

\institute{
 \begin{tabular}{cc}
         \begin{tabular}{c}
         $^1$  Microsoft Research\\
               7 J J Thomson Avenue\\
               Cambridge, United Kingdom\\
               \email{youssefh@microsoft.com}
         \end{tabular}
         & ~~
         \begin{tabular}{c}
         $^2$ Université Lille-Nord de France\\
              CRIL - CNRS UMR 8188\\
              Artois, F-62307 Lens\\
              \email{\{jabbour,sais\}@cril.fr}              
         \end{tabular}
 \end{tabular}
}

\maketitle

\begin{abstract}
This paper presents an original dynamic subsumption technique for Boolean CNF formulae.  It exploits simple and sufficient conditions to detect, during conflict analysis, clauses from the formula that can be reduced by subsumption. During the learnt clause derivation, and at each step of the associated resolution process, checks for backward subsumption between the current resolvent and clauses from the original formula are efficiently performed. The resulting method allows the dynamic removal of literals from the original clauses. Experimental results show that the integration of our dynamic subsumption technique within the state-of-the-art SAT solvers Minisat and Rsat particularly benefits to crafted problems.

\end{abstract}

\section{Introduction}
The SAT problem, i.e., the problem of checking whether a set of Boolean clauses is satisfiable or not, is central to many domains in computer science and artificial intelligence including constraint satisfaction problems (CSP), planning, non-monotonic reasoning, VLSI correctness checking, etc. Today, SAT has gained a considerable audience with the advent of a new generation of SAT solvers able to solve large instances encoding real-world applications and the demonstration that these solvers represent important low-level building blocks for many important fields, e.g., SMT solving, Theorem proving, Model finding, QBF solving, etc. These solvers, called modern SAT solvers \cite{Moskewicz01,MiniSat03}, are based on classical unit propagation \cite{Davis62} efficiently combined through incremental data structures with: (i) restart policies \cite{Gomes1998,kautz02dynamic}, (ii) activity-based variable selection heuristics (VSIDS-like) \cite{Moskewicz01}, and (iii) clause learning \cite{Bayardo97,Marques-Silva96,Moskewicz01}.  Modern SAT solvers can be seen as an extended version of the well known DPLL-like  procedure obtained thanks to these different enhancements. It is important to note that the well known resolution rule \`a la Robinson still plays a strong role in the efficiency of modern SAT solvers which can be understood as a particular form of general resolution \cite{BeameKS03}.

Indeed, conflict-based learning, one of the most important component of SAT solvers is based on resolution. We can also mention, that the well known and highly successful ({\tt SatElite})  preprocessor is based on variable elimination through the resolution rule \cite{SubbarayanP04a,Biere05}.  As mentioned in \cite{SubbarayanP04a}, on industrial instances, resolution leads to the generation of many tautological resolvents. This can be explained by the fact that many clauses represent Boolean functions encoded through a common set of variables. This property of the encodings might also be at the origin of many redundant or subsumed clauses at different steps of the search process.



The utility of ({\tt SatElite}) on industrial problems has been proved, and therefore one can wonder if the application of the resolution rule could be performed not only as a pre-processing stage but systematically during the search process. Unfortunately, dynamically maintaining a formula closed under subsumption might be time consuming. An attempt has been made recently in this direction by L. Zhang \cite{Zhang05}. In this work, a novel algorithm maintains a subsumption-free clause database by dynamically detecting and removing subsumed clauses as they are added. Interestingly, the author mention the following perspective of research:  "How to balance the runtime cost and the quality of the result for on-the-fly CNF simplification is a very interesting problem worth much further investigation".

In this paper, our objective is to design an effective dynamic simplification algorithm based on resolution. Our proposed approach aims at eliminating literals from the CNF formula by dynamically substituting smaller clauses.  More precisely, our approach exploits the intermediate steps of classical conflict analysis to subsume the clauses of the formula which are used in the underlying resolution derivation of the asserting clause. Since original clauses or learnt clauses can be used during conflict analysis both categories can be simplified. The effectiveness of our technique lies in the efficiency of the subsumption test, which is based on a simple and sufficient condition computable in constant time. Moreover, since our technique relies on the derivation of a conflict-clause, it is guided by the conflicts, and simplifies parts of the formula identified as important by the search strategy (VSIDS guidance). This dynamic process preserves the satisfiability of the formula, and with some additional bookkeeping can preserve the equivalence of the models.

The paper is organized as follows. After some preliminary definitions and notations, classical implication graph and learning schemes are presented in section \ref{sec:base}.  Then our dynamic subsumption approach is described in section \ref{sec:sub}. Finally, before the conclusion, experimental results demonstrating the performances of our approach are presented.

\section{Technical background}\label{sec:base}\label{sec:learning}

\subsection{Preliminary definitions and notations}
A {\it CNF formula} ${\cal F} $ is a
conjunction of {\it clauses}, where a clause is a disjunction of {\it literals}.  A literal is a positive ($x$) or
negated ($\neg x$)
propositional variable.  The two literals $x$ and $\neg x$ are called
{\it complementary}.
We note by $\bar{l}$  the complementary literal of $l$.
For a set of literals $L$, $\bar{L}$ is defined as $\{\bar{l} ~|~ l \in L\}$. 
A {\it unit} clause is a clause containing only one literal (called
{\it unit literal}), while a  binary clause contains exactly two
literals. An \emph{empty clause}, noted $\perp$, is interpreted as
false (unsatisfiable), whereas an \emph{empty CNF formula}, noted
$\top$, is interpreted as true (satisfiable).



The set of variables occurring in ${\cal F}$ is noted $V_ {\cal F}$.
A set of literals is \emph{complete} if it contains one literal for
each variable in $V_{\cal F}$,
and \emph{fundamental} if it does not contain complementary literals.
An {\it assignment} $\rho$ of a Boolean formula ${\cal F}$ is function which 
associates a value $\rho(x)\in\{false, true\}$
to some of the variables $x \in \cal F$.
$\rho$ is \emph{complete} if it assigns a value to every $x \in \cal
F$, and \emph{partial} otherwise.
An assignment is alternatively represented by a complete and
fundamental set of literals, in the
obvious way.
A {\it model} of a formula ${\cal F}$ is an  assignment $\rho$
that  makes the  formula $true$; noted $\rho\vDash\Sigma$.

The following notations will be heavily used throughout the paper:
\begin{itemize}
\item  $\eta[x, c_i, c_j]$ denotes the \emph{resolvent} between a clause $c_i$
containing the literal $x$ and $c_j$ a clause containing the opposite
literal $\neg x$.
In other words $\eta[x, c_i, c_j] = c_i\cup c_j\backslash \{x, \neg x\}$.
A resolvent is called {\it tautological} when it contains opposite literals.

\item ${\cal F}|_x$ will denote the formula obtained from ${\cal F}$
by assigning $x$ the truth-value \true.
Formally ${\cal F}|_x = \{c ~|~ c\in {\cal F}, \{x, \neg x\} \cap c =
\emptyset\} \cup
\{c\backslash \{\neg x\} ~|~ c\in {\cal F},  \neg x\in c\}$
(that is: the clauses containing $x$ and are therefore satisfied are removed;
and those containing $\neg x$ are simplified).
This notation is extended to assignments: given an assignment
$\rho=\{x_1,\dots, x_n\}$,
we define ${\cal F}|_\rho= (\dots (({\cal F}|_{x_1})|_{x_2})\dots |_{x_n})$.

\item ${\cal F}^*$ denotes the formula ${\cal F}$ closed under unit
propagation, defined recursively as follows:
(1) ${\cal F}^* = {\cal F}$ if ${\cal F}$ does not contain any unit clause,
(2)  ${\cal F}^* = \perp$ if ${\cal F}$ contains two unit-clauses
$\{x\}$ and  $\{\neg x\}$,
(3) otherwise, ${\cal F}^*= ({\cal F}|_{x})^*$ where $x$ is the
literal appearing in a unit clause of ${\cal F}$.  
A clause $c$ is deduced by unit propagation from ${\cal F}$, noted
${\cal F}\models^{*} c$, if  $({\cal F}|_{\bar{c}})^*=\perp$.

\end{itemize}
Let $c_1$ and $c_2$ be two clauses of a formula ${\cal F}$. We say that $c_1$ (respectively $c_2$) 
subsume (respectively is subsumed) $c_2$ (respectively by $c_1$) iff  $c_1 \subseteq c_2$. If $c_1$ 
subsume $c_2$, then   $c_1 \models c_2$ (the converse is not true). Also ${\cal F}$ and ${\cal F} -{c_2}$ 
are equivalent with respect to satisfiability. 


\subsection{DPLL search}
DPLL \cite{Davis62} is a tree-based backtrack search procedure; at each node of the search tree,
the assigned literals (decision literal and the propagated ones) are
labeled with the same \emph{decision level} starting from 1 and
increased at each decision (or branching).  After
backtracking, some variables are unassigned, and the current decision
level is decreased accordingly.  At level $i$, the current partial
assignment $\rho$ can be represented as a sequence of
decision-propagation of the form $\langle( x_k^i),x_{k_1}^i,
x_{k_2}^i,\dots, x_{k_{n_k}}^i \rangle$ where the first literal
$x_k^i$ corresponds to the decision literal $x_k$ assigned at level
$i$ and each $x_{k_j}^i$ for $1\leq j\leq n_k$ represents a propagated
(unit) literals at level $i$.  Let $x\in\rho$, we note $l(x)$ the
assignment level of $x$. For a clause $\alpha$, $l(\alpha)$ is
defined as the maximum level of its assigned literals.  

\subsection{Conflict analysis using implication graphs}
\label{sec:confl}
Implication graphs is a standard representation conveniently
used to analyze conflicts in modern SAT solvers. Whenever a literal $y$ is propagated,
 we keep a reference to the clause which triggers the propagation
of $y$, which we note $\classicalclause(y)$.
 The clause $\classicalclause(y)$, called implication of $y$, is in
this case of the form $(x_1 \vee \dots \vee x_n \vee y)$
 where every literal $x_i$ is false under the current partial assignment
 ($\rho(x_i) = \false, \forall i \in 1..n$), while $\rho(y) = \true$.
 When a literal $y$ is not obtained by propagation but comes from a
decision, $\classicalclause(y)$ is
 undefined, which we note for convenience $\classicalclause(y) = \perp$.  When 
 $\classicalclause(y) \not= \perp$, we denote by $exp(y)$ the set
 $\{\overline{x} ~|~ x \in \classicalclause(y)\setminus \{y\}\}$,
called set of \emph{explanations} of $y$.  When $\classicalclause(y)$ is undefined we 
define $exp(y)$ as the empty set.

 \begin{definition}[Implication Graph]
 Let ${\cal F}$ be a CNF formula, $\rho$ a partial assignment,
 and let $exp$ denotes the set of explanations for the deduced (unit propagated) 
literals in $\rho$.
 The implication graph associated to ${\cal F}$, $\rho$ and $exp$ is
 ${\cal G}_{\cal F}^{\rho, exp}=({\cal N},{\cal E})$ where:

 \begin{itemize}
 \item ${\cal N} = \rho$, i.e., there is exactly one node for every
literal, decided or implied;
 \item ${\cal E} = \{(x, y)  ~|~  x \in \rho, y \in \rho, x \in
\textsl{exp}(y)\}$
 \end{itemize}
 \end{definition}

In the rest of this paper,  for simplicity reason, $exp$ is omitted,
and an implication   graph is simply  noted as ${\cal G}_{\cal
F}^{\rho}$.  We also note $m$ as the conflict level.

\begin{example}
 \label{ex:formula1}
 ${\cal G}^\rho_{\cal F}$, shown in Figure \ref{fig:ig-cuts} is an
implication graph for
 the formula ${\cal F}$ and the partial assignment $\rho$ given below
: ${\cal F} \supseteq \{c_1,\dots,c_{12}\}$
 \[
 \begin{array}{llrlrl}
   (c_1) & \neg x_1    \vee \neg x_{11} \vee x_{2}  ~~&
   (c_2) & \neg x_1 \vee x_{3} ~~&
   (c_3) & \neg x_2    \vee  \neg x_{12} \vee x_4   \\
   (c_4) & \neg x_1    \vee  \vee  \neg x_3 \vee x_5 ~~&
   (c_5) & \neg x_4    \vee \neg x_5 \vee \neg x_6 \vee x_7 ~~ &
   (c_6) & \neg x_5    \vee \neg x_6 \vee x_8\\
   (c_7) & \neg x_7    \vee x_9 ~~&
   (c_8) & \neg x_5     \vee \neg x_8 \vee  \neg x_9 ~~ &
   (c_{9}) & \neg x_{10}  \vee \neg x_{17}   \vee x_1 ~~\\
   (c_{10}) & \neg x_{13}     \vee \neg x_{14} \vee  x_{10} ~~ &
   (c_{11}) &\neg x_{13} \vee  x_{17} ~~ &
   (c_{12}) &\neg x_{15} \vee \neg x_{16} \vee  x_{13}  \\

 \end{array}
 \]

\begin{figure}[htbp]
\centering
\includegraphics[width=9.5cm]{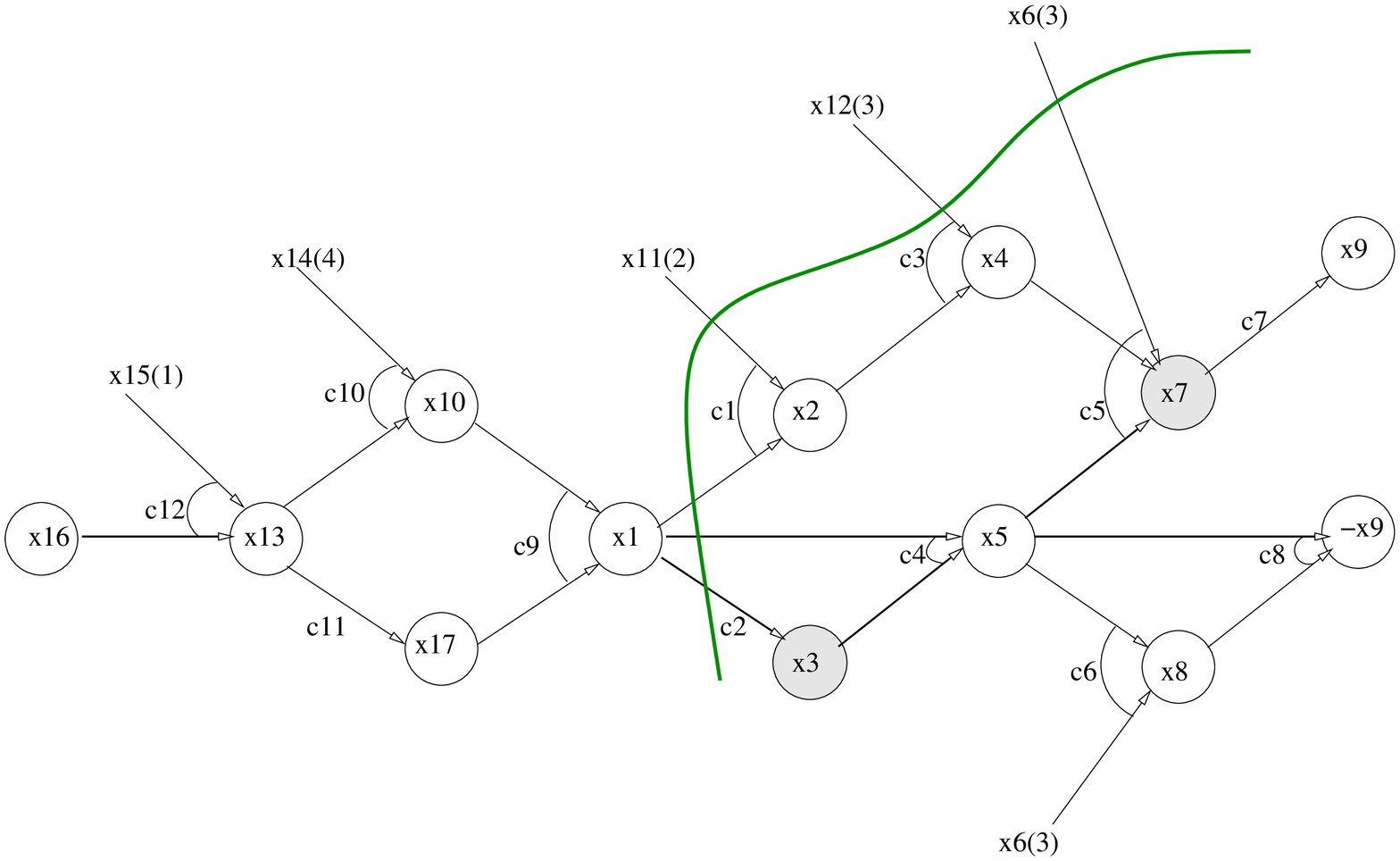}
  \caption{Implication Graph ${\cal G}_{\cal F}^\rho =  ({\cal N},{\cal E})$}
  \label{fig:ig-cuts}
\end{figure}

\noindent
$\rho=\{\langle\dots \neg x_{15}^1 \ldots\rangle
\langle(x_{11}^2)\dots \dots\rangle \langle(x_{12}^3)\dots
x_{6}^3\dots\rangle \langle(x_{14}^4),\dots \rangle \langle(x_{16}^5),x_{13}^5,\dots \rangle\}$. 
The conflict level is $5$ and $\rho({\cal F}) = false$.\\

\end{example}
%

%

%
\begin{definition}[Asserting clause]
  A clause $c$ of the form $(\alpha\vee x)$ is called an asserting clause iff 
  $\rho(c)=\false$, $l(x) = m$ and $\forall y\in\alpha, l(y)<l(x)$. 
  $x$ is called asserting literal. 
\end{definition}

Conflict analysis is the result of the application of resolution starting from the conflict clause using different implications implicitly encoded in the implication graph. We call this  process an asserting clause derivation (in short ACD).

\begin{definition}[Asserting clause derivation]
\label{def:crp}
An asserting clause derivation $\pi$ is a sequence of clauses 
$\langle \resolvent_1,\resolvent_2,\dots \resolvent_k\rangle$ 
satisfying the following conditions :

\begin{enumerate}

\item $\resolvent_1 = \eta[x, \classicalclause(x), \classicalclause(\neg x)]$, 
  where $\{x, \neg x\}$ is the conflict.

\item $\resolvent_i$, for $i  \in 2..k$, is built by selecting a literal $y \in \resolvent_{i-1}$ 
  for which $\classicalclause(\overline{y})$ is defined. We then have $y \in \resolvent_{i-1}$ and 
  $\overline{y} \in \classicalclause(\overline{y})$: the two clauses resolve. The clause $\resolvent_i$ is
  defined as $\eta[y, \resolvent_{i-1}, \classicalclause(\overline{y})]$;
  
\item $\resolvent_k$ is, moreover an asserting clause.

\end{enumerate}

\end{definition}

Note that every $\resolvent_i$ is a resolvent of the formula $\cal F$: by induction, $\resolvent_1$ is the resolvent
between two clauses that directly belong to $\cal F$; for every $i>1$, $\resolvent_i$ is a
resolvent between $\resolvent_{i-1}$ (which, by induction hypothesis, is a resolvent) and a clause of $\cal F$.
Every $\resolvent_i$ is therefore also an \emph{implicate} of $\cal F$, that is: ${\cal F} \models \resolvent_i$.  
By definition of the implication graph, we also have ${\cal F'} \models^* \resolvent_i$ where ${\cal F'}\subset {\cal F}$ 
is the set of clauses used to derive $\resolvent_i$.


%
Let us consider again the example \ref{ex:formula1}. The traversal of the graph ${\cal G}_{\cal F}^\rho$ (see Fig. \ref{fig:ig-cuts}) leads to the following asserting clause derivation: 
$\langle \resolvent_1,\dots, \resolvent_7 \rangle$ where $\resolvent_1 =  \eta[x_{9}, c_7, c_8] = (\neg{x_5}^5 \vee \neg{x_7}^5 \vee \neg{x_8}^3)$ and $ \resolvent_7 = (\neg{x_{11}}^2 \vee \neg{x_{12}}^3 \vee  \neg{x_{6}}^3 \vee \neg{x_1}^5)$. The node $x_1$ corresponding to the asserting literal $\neg x_1$ is called the first Unique Implication Point (First UIP).  





\section{Learning for dynamic subsumption}
\label{sec:sub}

In this section, we show how classical learning can be adapted for an efficient dynamic subsumption of clauses.  In our approach, we exploit the intermediate steps or resolvents generated during the classical conflict analysis to subsume some of the clauses used in the underlying resolution derivation of the asserting clause.  Obviously, it would be possible to consider the subsumption test between each generated resolvent and the whole set of clauses. However, this could be very costly in practice. Let us, illustrate some of the main features of our proposed approach. 

\subsection{Motivating example}
Let us reconsider again the example \ref{ex:formula1} and the implication graph ${\cal G}_{\cal F}^\rho$ (figure \ref{fig:ig-cuts}).  The asserting clause derivation  leading to the asserting clause $\Delta_1$  is described as follows: \\
$\pi = \langle \resolvent_1,\resolvent_2, \resolvent_3, \dots, \resolvent_7 = \Delta_1\rangle$

 \begin{itemize}
\item $\sigma_1 = \eta[x_{9}, c_7, c_8] = (\neg{x_8}^5 \vee \neg{x_7}^5 \vee \neg{x_5}^5)$
\item $\sigma_2 = \eta[x_{8}, \sigma_1, c_6] =  (\neg{x_6}^3 \vee  \neg{x_7}^5 \vee \neg{x_5}^5)$
\item $\sigma_3 = \eta[x_{7}, \sigma_2, c_5] = (\neg{x_6}^3 \vee \neg{x_5}^5\vee \neg{x_4}^5)\subset c_5$ (subsumption) 
 \item $\dots$
 \item $ \Delta_1 = \sigma_7 = \eta[x_2, \sigma_6, c_1] = 
(\neg{x_{11}}^2 \vee \neg{x_{12}}^3 \vee  \neg{x_{6}}^3 \vee \neg{x_1}^5)$
   
 \end{itemize}

As we can see the asserting clause derivation $\pi$ includes the resolvent  
$\sigma_3 = (\neg{x_6}^3 \vee \neg{x_5}^5\vee \neg{x_4}^5)$ which subsumes the clause 
$c_5 = (\neg x_6 \vee \neg x_5 \vee \neg x_4\vee x_7)$. Consequently, the literal $x_7$ 
is eliminated from the clause $c_5$.   In general, the resolvent $\sigma_3$ can subsume 
other clauses from the implication graph that include the literals $\neg x_6$, $\neg x_5$ and  $\neg x_4$. 

\subsection{Dynamic subsumption: a general formulation}
 \label{sec:genfor}
Let us now give a formal presentation of our dynamic subsumption approach. 

\begin{definition}[F-subsumption modulo UP]
Let $c\in{\cal F}$.
$c$ is ${\cal F}$-subsumed modulo Unit Propagation  iff $\exists c'\subset c$ such that ${\cal F}|_{\bar{c'}} \models^* \bot$
\end{definition}

Given two clauses $c_1$ and $c_2$ from ${\cal F}$ such that $c_1$ subsumes $c_2$, then $c_2$ is ${\cal F}$-subsumed modulo UP. 

As explained before, subsuming clauses during search might be time consuming. In our proposed framework, to reduce the computational cost, we restrict subsumption checks to the intermediate resolvents $\resolvent_i$ and the clauses of the form $\classicalclause({y})$ used to derive them (clauses encoded in the implication graph).

\begin{definition}
Let ${\cal F}$ be a formula and $\pi = \langle\resolvent_1 \ldots \resolvent_k\rangle$ an asserting clause derivation.
For each $\resolvent_i \in \pi$, we define ${\cal C}_{\resolvent_i} = \{\classicalclause({y}) \in {\cal F}| \exists
j \leq i$ st.  $\resolvent_j = \eta[y, \classicalclause({y}), \resolvent_{j-1}]\}$   as the set of clauses of ${\cal F}$ used for the derivation of $\resolvent_i$.
\end{definition}


\begin{property}
\label{pro:subclauses}
Let ${\cal F}$ be a formula and $\pi =  \langle \resolvent_1,\resolvent_2,\dots, \resolvent_i,\dots, \resolvent_k\rangle$ an asserting clause derivation. 
If $\resolvent_i$ subsumes a clause $c$ of ${\cal C}_{\resolvent_k}$ then $c\in {\cal C}_{\resolvent_i}$.
\end{property}
\begin{proof}
As $\resolvent_{i+1} = \eta[y, \classicalclause({y}), \sigma_i]$ where $\neg y\in \sigma_i$, we have $\resolvent_i\not\subset \classicalclause({y})$. The next resolution steps can not involve clauses containing the literal $\neg y$. Otherwise, the literal $y$ in the implication graph will admit two possible explanations (implications), which is not possible by definition of the implication graph. Consequently, $\sigma_i$ can not subsume clauses from ${\cal C}_{\resolvent_k}- {\cal C}_{\resolvent_i}$.
\end{proof}

\begin{property}
\label{pro:unitsub}
Let ${\cal F}$ be a formula and $\pi$ an asserting clause derivation. If $\resolvent_i\in\pi$ subsumes a clause 
$c$ of ${\cal C}_{\resolvent_i}$ then $c$ is ${\cal C}_{\resolvent_i}$-subsumed modulo UP.
\end{property}
\begin{proof}
As $\resolvent_i\in\pi$ is derived from  ${\cal C}_{\resolvent_i}$ by resolution, then  ${\cal C}_{\resolvent_i}\models \resolvent_i$.   By definition of an asserting clause derivation and implication graphs, we also have ${\cal C}_{\resolvent_i} \models^* \resolvent_i$ (see section \ref{sec:confl}).  As  $\resolvent_i$ subsumes $c$ ($\resolvent_i\subseteq c$), then ${\cal C}_{\resolvent_i}\models^* c$. 
\end{proof}

The Property \ref{pro:unitsub} shows that if a clause $c$ encoded in the implication graph is subsumed by $\resolvent_i$, such subsumption can be captured by subsumption modulo UP, while the Property \ref{pro:subclauses} mention that subsumption checks of  $\resolvent_i$ can be restricted to clauses from ${\cal C}_{\resolvent_i}$.  Consequently, a possible general dynamic subsumption approach can be stated as follows:
Let $\pi = \langle\resolvent_1,\ldots,\resolvent_i,\dots, \resolvent_k\rangle$ be an asserting resolution derivation.  For each resolvent $\resolvent_i\in\pi$, we apply subsumption checks between $\resolvent_i$  and all the clauses in ${\cal C}_{\resolvent_i}$. 

In the following, we show that we can reduce further the number of clauses to be checked for subsumption by considering only a subset of   ${\cal C}_{\resolvent_i}$. Obviously,  as $\resolvent_i$ is a resolvent of an asserting clause derivation $\pi$, then there exists two paths from the conflict nodes $x$ and $\neg x$ respectively,  to one or more nodes of the implication graph associated to the literals of $\resolvent_i$ assigned at the conflict level.  Consequently, we derive the following property:


\begin{property}
\label{pro:subsetCl}
Let $\pi$ be an asserting clause derivation, $\resolvent_i\in\pi$ and $c \in { \cal C}_{\resolvent_i}$. If $\resolvent_i$ subsumes $c$, then  there exists two paths from the conflict nodes $x$ and $\neg x$ respectively,  to one or more nodes of the implication graph associated to the literals of $c$ assigned at the conflict level.
\end{property}

The proof of the property is immediate since $\resolvent_i\subset c$. As this property is true for $\resolvent_i$ which is derived by resolution from the two clauses involving $x$ and $\neg x$. Then it is also, true for its supersets ($c$).

For a given $\resolvent_i$, the Property \ref{pro:subsetCl} leads us to another restriction of the set of clauses to be checked for subsumption. Indeed, we only need to consider the set of clauses ${ \cal P}_{\resolvent_i}$, linked (by paths) to the two conflicting literals $x$ and $\neg x$. \\

We illustrate this characterization using the example \ref{ex:formula1} (see. also Figure \ref{fig:ig-cuts}).  Let $\pi =\langle \resolvent_1, \dots, \resolvent_7 \rangle $ where $\resolvent_7 = (\neg{x_{11}} \vee \neg{x_{12}} \vee  \neg{x_{6}} \vee \neg{x_1})$. We have ${ \cal C}_{\resolvent_7} = \{c_1,c_2, c_3, c_4, c_5, c_6, c_7, c_8 \}$ and  ${ \cal P}_{\resolvent_7} = \{c_1, c_2, c_4, c_5, c_6, c_8 \}$. Indeed, from the  nodes $x_2$  of the clause $c_3$ we only have one path to the  nodes $x_9$. Consequently, the clause $c_3$ might be discarded from the set of clauses to be checked for subsumption. Similarly, the node $x_7$ of the clause $c_7$ is not linked with a path to $\neg x_9$. Then $c_7$ is not considered for subsumption tests.
 
\begin{property}
Given an asserting clause derivation $\pi =\langle \resolvent_1, \dots,\resolvent_k \rangle $. The time complexity of our general dynamic subsumption approach is in $O(|{ \cal C}_{\resolvent_k}|^2$).
\end{property}
\begin{proof}
From the definition of ${ \cal C}_{\resolvent_i}$, we have $|{ \cal C}_{\resolvent_i}| = i+1$. In the worst case, we need to consider $i+1$ subsumption checks. Then for all $\resolvent_i$ with $1\leq i\leq k$, we have to check $\sum_{1\leq i\leq k} (i+1) = \frac{k\times(k+3)}{2} $.  As $k= |{ \cal C}_{\resolvent_k}|$, then the worst case complexity is in $O(|{ \cal C}_{\resolvent_k}|^2$).
\end{proof}

The worst case complexity is quadratic even if we consider ${ \cal P}_{\resolvent_k}\subset { \cal C}_{\resolvent_k}$. 

\subsection{Dynamic subsumption on the fly}

In section \ref{sec:genfor}, we have presented the general approach  for dynamic subsumption. Its complexity is quadratic in the number of clauses used in the derivation of an asserting clause. As stated, in the introduction, to design an efficient dynamic simplification technique, one need to balance the run time cost and the quality of the simplification. In this section, we propose a restriction of the general dynamic subsumption scheme, called dynamic susbumption on the fly, which applies subsumption only between the current resolvent $\resolvent_i$ and the last clause from the implication graph used for its derivation.  More precisely, suppose $\resolvent_i =   \eta[y, c, \resolvent_{i-1}]$, we only check subsumption between $\resolvent_i$ and  $c$. 

The following  property gives a sufficient condition under which $y$ can be removed form $c$

\begin{property}
\label{pro:restrict}
 Let $\pi$ be an asserting clause derivation, $\resolvent_i\in\pi$ such that $\resolvent_i = \eta[y, c, \resolvent_{i-1}]$.
 If $\resolvent_{i-1}-\{y\} \subseteq c$, then $c$ is subsumed by $\resolvent_i$.
\end{property}

\begin{proof}
Let $c = (\neg y \vee \alpha)$ and  $\resolvent_{i-1} = (y \vee \beta)$. Then  $\resolvent_i=(\alpha \vee \beta)$. As $\resolvent_{i-1}-\{y\} \subseteq c$, then $\beta\subseteq\alpha$. So, $\resolvent_i = \alpha$ which subsumes  $(\neg y \vee \alpha) = c$.
\end{proof}

Considering modern SAT solvers that include conflict analysis, the integration of  this new dynamic subsumption approach can be done with negligible additional cost. Indeed, by using a simple counter during the conflict analysis procedure, 
we can  verify the sufficient condition given in the Property \ref{pro:restrict} with a constant complexity time. Indeed, at each step of the asserting clause derivation,  we generate the next resolvent $\resolvent_i$ from a clause $c$ and a resolvent $\resolvent_{i-1}$. In the classical implementation of conflict analysis, one can check in constant time if a given literal is present in the current resolvent. Consequently, during the visit of the clause $c$, we additionally compute the number $n$ of literals of $c$ that belong to $\resolvent_{i-1}$.  If $n \geq |\resolvent_{i-1}|-1$ then $c$ is subsumed by $\resolvent_i = \eta[y, c, \resolvent_{i-1}]$.

\section{Experiments}

The experiments were done on a large panel  of crafted and industrial problems coming from the last competitions. All the instances were simplified by the {\tt SatElite} preprocessor \cite{EenB05}. We implemented our dynamic subsumption approach in Minisat \cite{MiniSat03} and Rsat \cite{darwiche07} and made a comparison between the original solvers and the ones enhanced with dynamic subsumption. All the tests were made on a Xeon 3.2GHz (2 GB RAM) cluster. Results are reported in seconds.

\subsection{Crafted problems}

During these experiments, the CPU time limit was fixed to 3 hours. These problems are hand made and many of them are designed to beat all the existing DPLL solvers. They contain for example Quasi-group instances, forced random SAT instances, counting, ordering and pebbling instances, social golfer problems, etc.

\begin{figure}[htbp]
  \hspace{-0.5cm}    \includegraphics[width=6.3cm]{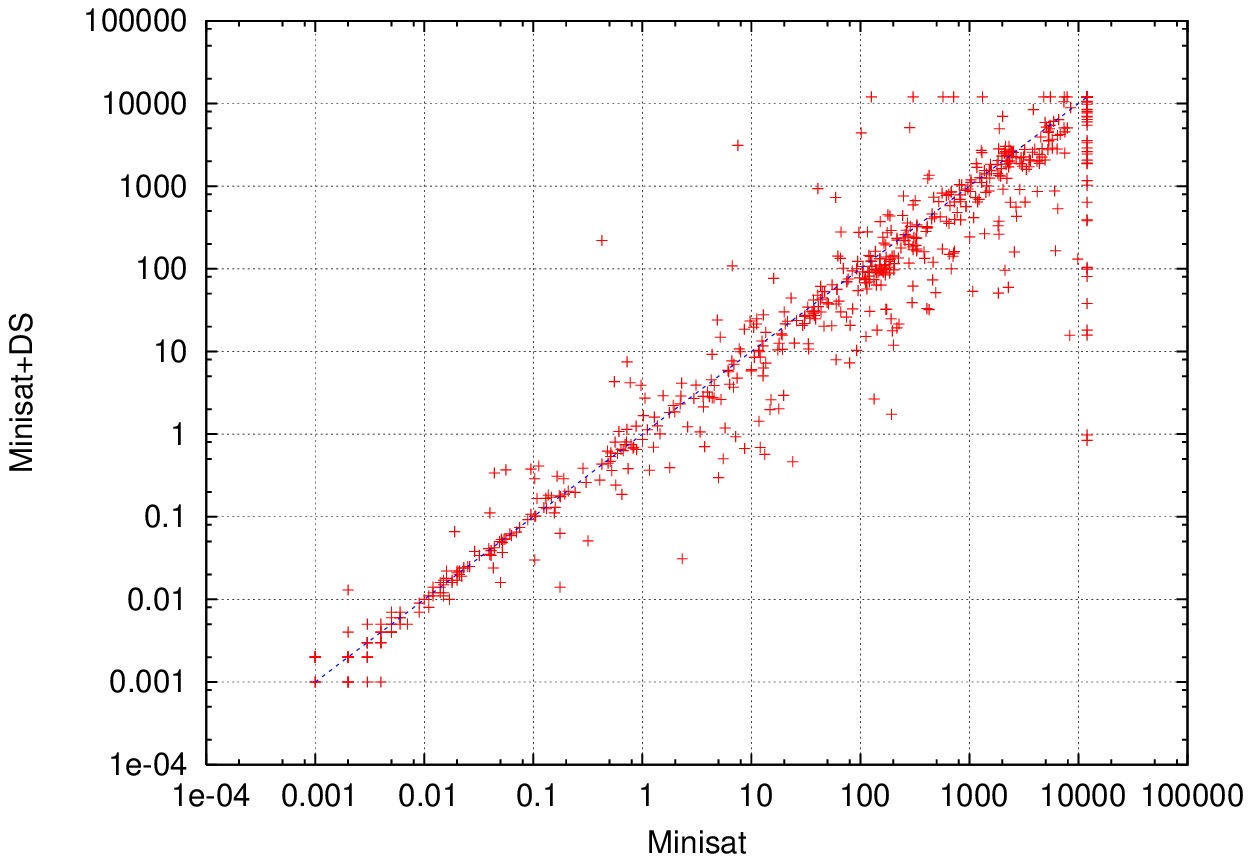}
\hspace{0.02cm}\includegraphics[width=6.3cm]{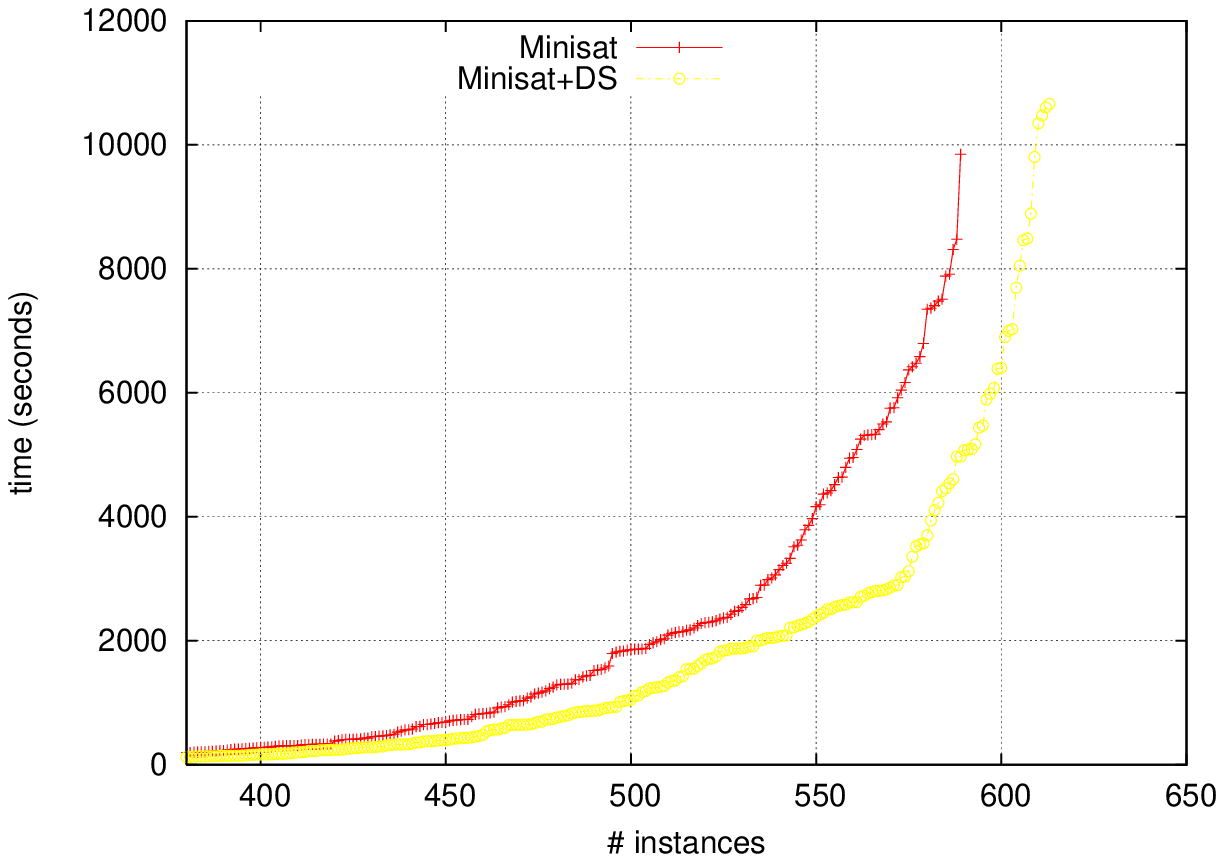}
\vspace{-0.3cm}
\caption{Crafted problems: \Minisat vs \Minisats}
\label{fig:minisat_c}
\end{figure}

The log-scaled scatter plot (in log scale) given in the left part of Figure \ref{fig:minisat_c} details the results for  \Minisat and \Minisats on each crafted instance. The x-axis (resp. y-axis) corresponds to the CPU time $tx$ (resp. $ty$) obtained by \Minisat (resp. \Minisats). Each dot with $(tx, ty)$ coordinates, corresponds to a SAT instance. Dots below (resp.  above) the diagonal indicate instances where the subsumption is more efficient i.e. $ty < tx$. This figure clearly shows the computational gain obtained thanks to our efficient dynamic subsumption approach. By automatically counting the points we found that 365 instances are solved more efficiently through dynamic subsumption. In some cases the gain is up to 1 order of magnitude. Of course, there exists instances where subsumption decreases the performances of \Minisat (178 instances).

The right part of the Figure \ref{fig:minisat_c} shows the same results with a different representation which gives for each technique the number of solved instances (\# instances) in less than $t$ seconds. This Figure confirms the efficiency of our dynamic subsumption approach on these problems. On several classes the number of removed literals is very important e.g., \texttt{x\_*}, \texttt{QG\_*}, \texttt{php\_*}, \texttt{parity\_*}. On the \texttt{genurq\_*}, \texttt{mod\_*}, and \texttt{urquhart\_*} the problem is simplified during each conflict analysis.


\begin{figure}[htbp]
  \hspace{-0.5cm}    \includegraphics[width=6.3cm]{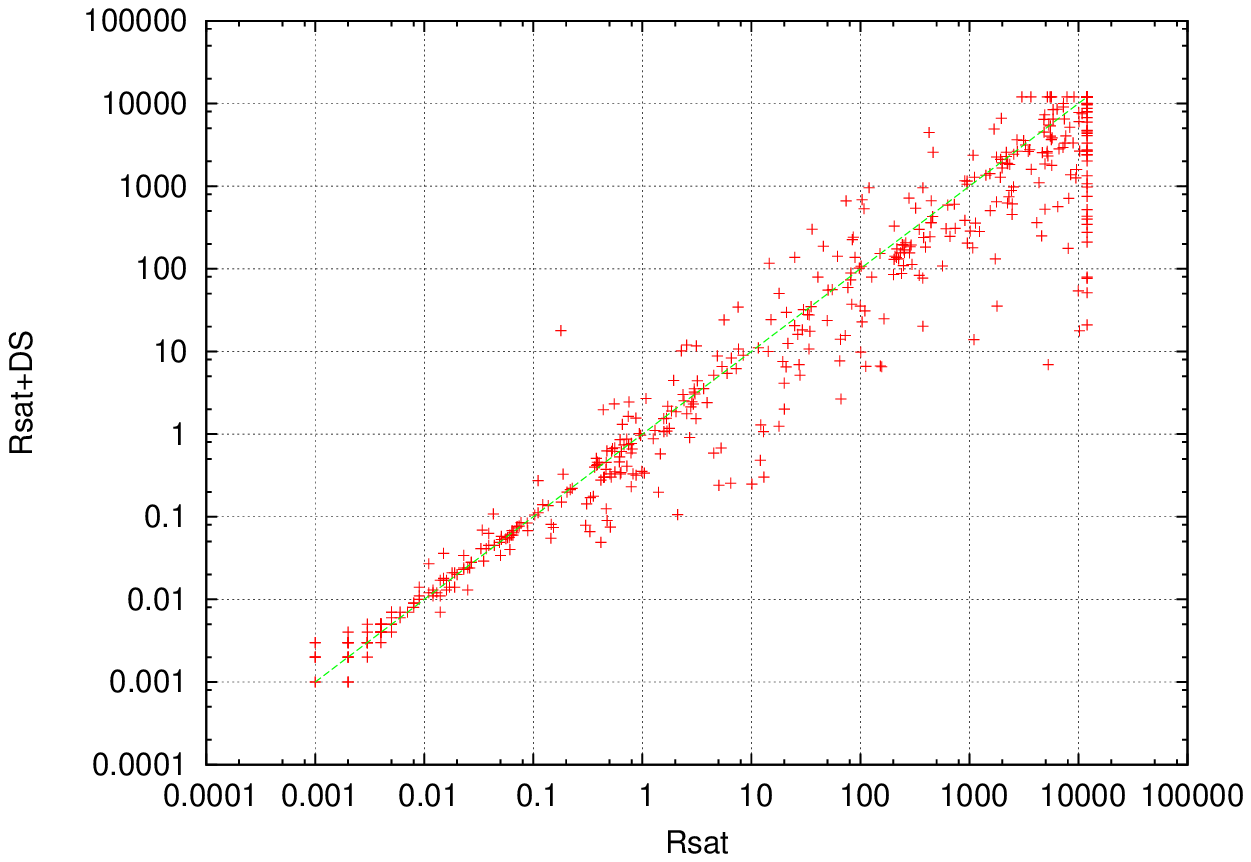}
\hspace{0.02cm}\includegraphics[width=6.3cm]{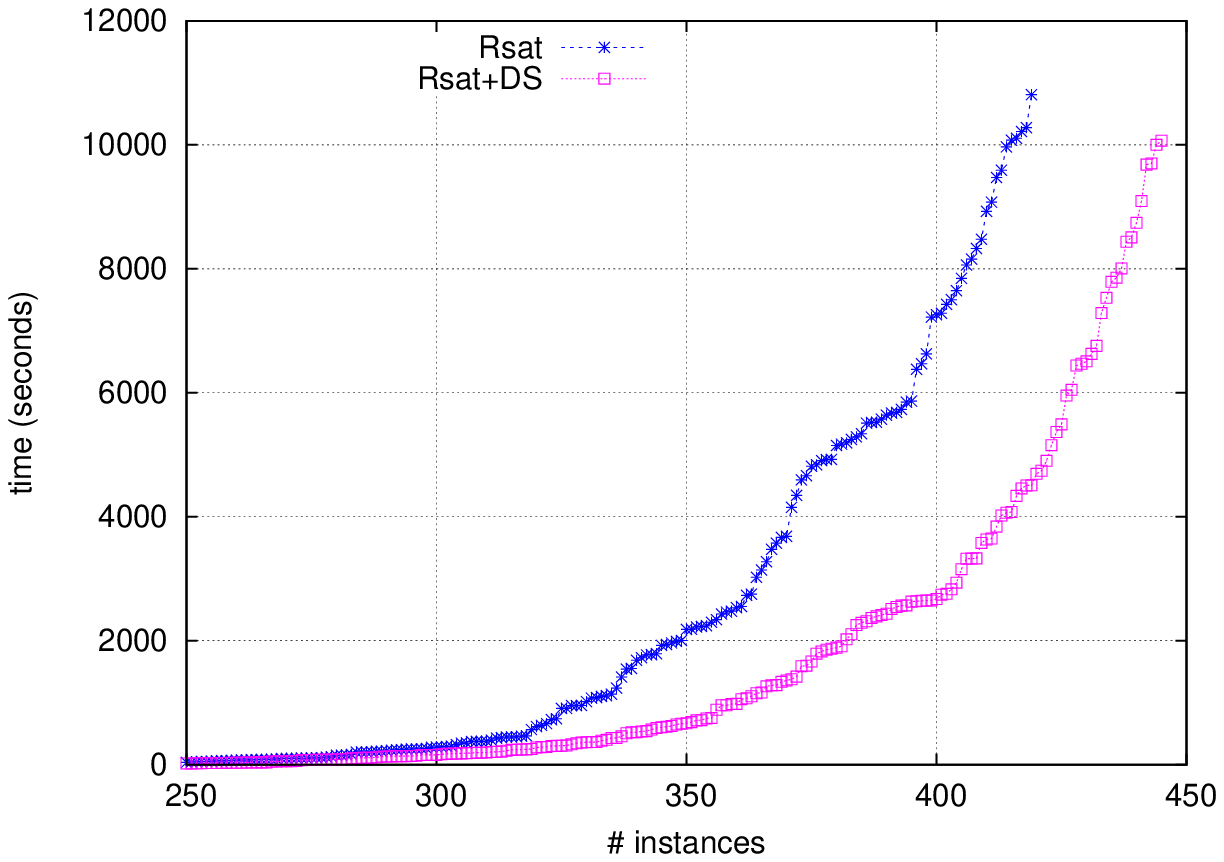}
\vspace{-0.3cm}
\caption{Crafted problems: \Rsat vs  \Rsats}

\label{fig:rsat_c}
\end{figure}

Figure \ref{fig:rsat_c} shows result for \Rsat and \Rsats. Overall we can see that the addition of our dynamic subsumption process to \Rsat improves the performance. The fine analysis of the left part of Figure \ref{fig:rsat_c} showed that \Rsats solves 327 instances more efficiently than \Rsat, which solves 219 problems more efficiently than its opponent.

Interestingly we can remark that the performances of \Rsat and \Rsats is worse than the ones of \Minisat and \Minisats. This comes from the rapid restart strategy used by this algorithm which does not pay off on crafted problems.

\subsection{Industrial problems}

With these problems, the time limit was set to 3 hours. Table \ref{tab:families} provides detailed results on the SAT industrial problems from the Sat-competition 2007 and Sat-Race 2008. The first column represents the instances families. The second column (\#inst.) indicates the total number of instances in each family. Following columns present results for respectively \Rsat, \Rsats, \Minisat, and \Minisats. In each of these columns the first number represents the number of instances solved, and the number in parenthesis represents the number of instances solved more quickly than the opponent. The last row of the table gives the total of each column. We can see that \Rsats  and \Minisats are in general faster and solves more problems than \Rsat  and \Minisat respectively.

\begin{table}[htbp]
  \centering
  \begin{small}
    \begin{tabular}{|l||l|l|l||l|l|}
      \hline
     {\bf  families} &{\bf  \# inst.} &{\bf \Rsat }    & {\bf \Rsats}    &  {\bf \Minisat }    & {\bf \Minisats}   \\
      \hline
      IBM\_*      & 53 & 15(7) & {\bf 17}({\bf 10})  & 15({\bf 10}) & 15(7)\\
      \hline
      APro\_*     & 16 & 12({\bf 7}) & 12(5) & {\bf 14}(6) & 13({\bf 8}) \\
      \hline
      mizh\_*     & 10 & 10({\bf 7}) & 10(3) & 10(5) & 10(5)  \\
      \hline
      Partial\_*  & 20 & 6(2) & {\bf 7}({\bf 5}) & 1(0) & {\bf 2}({\bf 2}) \\
      \hline
      total\_*    & 20 & 13(6) & 13({\bf 8})  & {\bf 10}(5) & 9({\bf 6}) \\
      \hline
      dated\_*    & 20 & 15(6) & {\bf 16}({\bf 10})  & {\bf 14}({\bf 10}) & 13(4) \\
      \hline
      braun\_*    &    7 & 4(1) & 4({\bf 3})  & 5(2) & 5({\bf 3}) \\
      \hline
      velev\_*    &10 & 2 (0) & 2({\bf 2}) & {\bf 2}({\bf 2}) & 1(0) \\
      \hline
      sort\_*     & 5 & 2({\bf 2}) & 2(0)&  2(0) & 2({\bf 2}) \\
      \hline
      manol\_*    & 10 & 8(3) & 8({\bf 5}) & 8({\bf 5}) & {\bf 9}(4) \\
      \hline
      vmpc\_*    & 10 & 9(1) & 9({\bf 8})  & 6(2) & {\bf 7}({\bf 5}) \\
      \hline
      clause\_*   & 5 & 3({\bf 2}) & 3(1)  & 3(0) & 3({\bf 3}) \\
      \hline
      cube\_*     & 4 & 4(2) & 4(2)  & 4(1) & 4({\bf 3}) \\
      \hline
      gold\_*     &4 &2 (2) & 2 ({\bf 0}) &  2(0) & 2({\bf 2}) \\
      \hline
      safe\_*     & 4 & 2(0) & 2({\bf 2})  & 1(0) & 1({\bf 1}) \\
      \hline
      simon\_*    & 5 & 5 ({\bf 4}) & 5 (1) &  5({\bf 3}) & 5(2) \\
      \hline
      block\_*    & 2 & 2({\bf 2}) & 2(0)  & 2(0) & 2({\bf 2})  \\
      \hline
      dspam\_*    & 10 & 10(5) & 10(5)& 10(5) & 10(5)\\
      \hline
      schup\_*    & 3 & 3(2) & 3(1) & 3({\bf 2}) & 3(1) \\ 
            \hline
      post\_*    &10  & 8(3) & 8({\bf 5}) &  5(3) & {\bf 6}(3)  \\
            \hline
        ibm\_*    &20 & 20(6) & 20({\bf 14})& 19(6) & 19({\bf 13}) \\
      \hline
      \hline
      Total & 248 & 155(70) & {\bf 159}({\bf 90})&  141(67)& 141({\bf 81}) \\
      \hline
    \end{tabular}
  \end{small}
  
  \caption{Industrial problems}
  \label{tab:families}
\end{table}

\begin{table}[htbp]
  \centering
  \begin{small}
    \begin{tabular}{|l||l|l|l||l|}
      \hline
          {\bf instances}              & {\bf \Rsat }   & {\bf \Rsats} &  {\bf \Minisat }  & {\bf \Minisats}   \\
          
      \hline
      \hline
vmpc\_33            &  5540  &  \textbf{1562}   &      --    &  --   \\
vmpc\_29            &  2598  &  \textbf{1302}    &      --    &  \textbf{1252}  \\
vmpc\_30            &  366  &  \textbf{105}   &      3111  &  \textbf{2039} \\
vmpc\_27            &  593  &  \textbf{327}   &      1159  &  \textbf{637} \\
vmpc\_31            &    --  &  --     &      --    &  -- \\
vmpc\_25            &  39    &  \textbf{1}  &      830   &  \textbf{318} \\
vmpc\_26            &  182  &  \textbf{69}   &      1239  &  \textbf{1235} \\
vmpc\_34            &  3366  &  \textbf{944}   &      --    &  -- \\
vmpc\_24            &  43 &  \textbf{8}   &      \textbf{82}  &  210 \\
vmpc\_28            &  \textbf{173}  & 488   &      \textbf{3859}   &  5478 \\
\hline
partial-5-11-s            &   931 & \textbf{176}     &      --       &  \textbf{2498} \\
partial-5-13-s            &   503 & \textbf{71}    &      3248.38  &  \textbf{669}\\
partial-5-15-s            &   \textbf{737.064} & 825    &      --       &   -- \\
partial-10-11-s           &   1242 & \textbf{875}     &      --       &   -- \\
partial-5-19-s            &   1134 & \textbf{498}    &      --       &   -- \\
partial-5-17-s            &   \textbf{7437.82} & 10610    &      --       &   -- \\
partial-10-13-s           &   --      & \textbf{3237}    &       --      &   -- \\
\hline                                                   
ibm-2002-04r-k80    & 90     & \textbf{33}   & \textbf{113}   & 152  \\
ibm-2002-11r1-k45   & 67     & \textbf{29 }  & 102   & \textbf{65}   \\
ibm-2002-18r-k90    & 265    & \textbf{157}  & 1044  & \textbf{769}  \\
ibm-2002-20r-k75    & \textbf{36}     & 185  & 2112  & \textbf{668}  \\
ibm-2002-22r-k60    & 738    & \textbf{691}  & 5480  & \textbf{3434} \\
ibm-2002-22r-k75    & 363    & \textbf{349}  & 1109  & \textbf{688}  \\
ibm-2002-22r-k80    & \textbf{285}    & 298  & 894   & \textbf{642}  \\
ibm-2002-23r-k90    & 1477   & \textbf{965}  & 7127  & \textbf{2670} \\
ibm-2002-24r3-k100  & 273    & \textbf{256}  & \textbf{133}   & 249  \\
ibm-2002-25r-k10    & \textbf{3104}   & 3118 & \textbf{2877}  & 3172 \\
ibm-2002-29r-k75    & 353    & \textbf{248}  & \textbf{272}   & 1107 \\
ibm-2002-30r-k85    & 3853   & \textbf{592}  & -- & --   \\
ibm-2002-31\_1r3-k30 & 1203   & \textbf{652}  & 1150  & \textbf{998}  \\
ibm-2004-01-k90     &  114   & \textbf{30}   & \textbf{251}   & 726  \\
ibm-2004-1\_11-k80   & 394    & \textbf{222}  & 559   & \textbf{329}  \\
ibm-2004-23-k100    & \textbf{326}    & 687  & 3444  & \textbf{2743} \\
ibm-2004-23-k80     &  \textbf{465}   & 563  & 2060  & \textbf{1584} \\
ibm-2004-29-k25     &  290   & \textbf{210}  & 1061  & \textbf{1017} \\
ibm-2004-29-k55     &  533   & \textbf{16 }  & 558   & \textbf{124}  \\
ibm-2004-3\_02\_3-k95 &  \textbf{1}     & 2    & \textbf{1}     & 2    \\
\hline
\end{tabular}
  \end{small}
  \caption{Zoom on industrial families}
  \label{tab:focus}
\end{table}

Table \ref{tab:focus}, focuses on some industrial families. In these families, the speed-up are relatively important. For instance, if we consider the vmpc family, we can see that our dynamic simplification allows a one order of magnitude improvement with \Rsats (instances 24, and 25). On the same family, on the 9 solved instances by \Rsats and \Rsat, \Rsats  is better on 8 instances. While \Minisats is better than \Minisat on 5 instances among the 7 solved instances.

Overall, our experiments allow us to demonstrate two things. First our technique does not degrade and often improve the performance of DPLLs on industrial problems. Second, it enhances the applicability of these algorithms on classes of problems which are made to be challenging for them. Since the implementing of our algorithm is rather simple, we think that overall, it represents an interesting contribution for the robustness of modern DPLLs.

\section{Related Works}
\label{sec:relatedWorks}

In Darras et al. \cite{DarrasDDMOS05}, the authors proposed a preprocessing based on unit propagation for sub-clauses deduction.  Considering the implication graph generated by the constraint propagation process as a resolution tree,  the proposed approach deduces sub-clauses from the original formula.  However, their proposed dynamic version is clearly time consuming. The experimental evaluation is only given in term of number of nodes.

In \cite{Zhang05}, an algorithm for maintaining a subsumption-free CNF clause database is presented. It efficiently detects and removes subsumption when a learnt clause is added. Additionally, the algorithm compacts the database greedily by recursively applying resolutions in order to decrement the size of the database. 

Conflict-clause shrinking was introduced in Minisat 1.14 \cite{minisat114}. It is also implemented in PicoSAT \cite{picosat}. It removes literals from learnt clauses by resolving recursively with clauses of the implication graph. Remark that in the previous experiments, our base solvers \Minisat and \Rsat implement this technique.

\section{Conclusion}
This paper presents a new subsumption technique for Boolean CNF formulae. It makes an original use of learning to reduce original or learnt clauses. At each conflict, and during the asserting clause derivation process, subsumption between the generated resolvents and some clauses encoded in the implication graph is checked using an efficient sufficient condition. Interestingly, since our subsumption technique relies on the clauses used in the derivation of an asserting clause, it tends to simplify parts of the formula identified as important by the activity-based search strategy.

Experimental results show that the integration of our method within two state-of-the-art SAT solvers \Minisat and \Rsat particularly benefits to crafted problems and achieves interesting improvements on several industrial families.

As a future work, we plan to investigate how to efficiently extend our approach to achieve exhaustive clauses subsumption. Another interesting path of research would be to exploit our subsumption framework to fine tune the activity based strategy. Indeed, each time a literal is eliminated, this mean that a new conflict clause is derived and all the resolvents used in such derivation are useless and can be dropped from the implication graph.

\bibliographystyle{plain}
\bibliography{subDynV16}

\begin{thebibliography}{10}

\bibitem{Bayardo97}
Roberto~J. {Bayardo, Jr.} and Robert~C. Schrag.
\newblock Using {CSP} look-back techniques to solve real-world {SAT} instances.
\newblock In {\em Proceedings of the Fourteenth National Conference on
  Artificial Intelligence (AAAI'97)}, pages 203--208, 1997.

\bibitem{BeameKS03}
Paul Beame, Henry~A. Kautz, and Ashish Sabharwal.
\newblock Understanding the power of clause learning.
\newblock In Georg Gottlob and Toby Walsh, editors, {\em IJCAI}, pages
  1194--1201. Morgan Kaufmann, 2003.

\bibitem{Biere05}
A~Biere and N.~Eén.
\newblock Effective preprocessing in sat through variable and clause
  elimination.
\newblock In {\em Proceedings of the Eighth International Conference on Theory
  and Applications of Satisfiability Testing (SAT'05)}, 2005.

\bibitem{picosat}
Armin Biere.
\newblock Picosat essentials.
\newblock {\em Journal on Satisfiability, Boolean Modeling and Computation
  (JSAT)}, 4(1):75--97, 2008.

\bibitem{DarrasDDMOS05}
S.~Darras, G.~Dequen, L.~Devendeville, B.~Mazure, R.~Ostrowski, and
  L.~Sa{\"i}s.
\newblock Using \protect{B}oolean constraint propagation for sub-clauses
  deduction.
\newblock In {\em Proceedings of the Eleventh International Conference on
  Principles and Practice of Constraint Programming(CP'05)}, pages 757--761,
  2005.

\bibitem{Davis62}
M.~Davis, G.~Logemann, and D.~W. Loveland.
\newblock A machine program for theorem-proving.
\newblock {\em Communications of the ACM}, 5(7):394--397, 1962.

\bibitem{minisat114}
N.~Een and N.~S\"orensson.
\newblock Minisat - a sat solver with conflict-clause minimization.
\newblock In {\em Proceedings of the Eighth International Conference on Theory
  and Applications of Satisfiability Testing (SAT'05)}, 2005.

\bibitem{EenB05}
N.~Eén and A.~Biere.
\newblock Effective preprocessing in \protect{SAT} through variable and clause
  elimination.
\newblock In {\em Proceedings of the Eighth International Conference on Theory
  and Applications of Satisfiability Testing (SAT'05)}, pages 61--75, 2005.

\bibitem{MiniSat03}
Niklas Eén and Niklas Sörensson.
\newblock An extensible sat-solver.
\newblock In {\em Proceedings of the Sixth International Conference on Theory
  and Applications of Satisfiability Testing (SAT'03)}, pages 502--518, 2002.

\bibitem{Gomes1998}
Carla~P. Gomes, Bart Selman, and Henry Kautz.
\newblock Boosting combinatorial search through randomization.
\newblock In {\em Proceedings of the Fifteenth National Conference on
  Artificial Intelligence (AAAI'97)}, pages 431--437, Madison, Wisconsin, 1998.

\bibitem{kautz02dynamic}
H.~Kautz, E.~Horvitz, Y.~Ruan, C.~Gomes, and B.~Selman.
\newblock Dynamic restart policies.
\newblock In {\em Proceedings of the Eighteenth National Conference on
  Artificial Intelligence (AAAI'02)}, pages 674--682, 2002.

\bibitem{Marques-Silva96}
Joao~P. Marques-Silva and Karem~A. Sakallah.
\newblock {GRASP - A New Search Algorithm for Satisfiability}.
\newblock In {\em Proceedings of IEEE/ACM International Conference on
  Computer-Aided Design}, pages 220--227, November 1996.

\bibitem{Moskewicz01}
M.~W. Moskewicz, C.~F. Madigan, Y.~Zhao, L.~Zhang, and S.~Malik.
\newblock Chaff: Engineering an efficient \protect{SAT} solver.
\newblock In {\em Proceedings of the 38th Design Automation Conference
  ({DAC}'01)}, pages 530--535, 2001.

\bibitem{darwiche07}
Knot Pipatsrisawat and Adnan Darwiche.
\newblock A lightweight component caching scheme for satisfiability solvers.
\newblock In {\em Proceedings of 10th International Conference on Theory and
  Applications of Satisfiability Testing(SAT)}, pages 294--299, 2007.

\bibitem{SubbarayanP04a}
Sathiamoorthy Subbarayan and Dhiraj~K. Pradhan.
\newblock \protect{NiVER}: Non-increasing variable elimination resolution for
  preprocessing \protect{SAT} instances.
\newblock In {\em Proceedings of the Seventh International Conference on Theory
  and Applications of Satisfiability Testing (SAT'04)}, pages 276--291, 2004.

\bibitem{Zhang05}
Lintao Zhang.
\newblock On subsumption removal and on-the-fly cnf simplification.
\newblock In {\em SAT'2005}, pages 482--489, 2005.

\end{thebibliography}
\end{document}